\documentclass[cameraready]{Interspeech}

\newif\ifanonymized
\anonymizedfalse
\newcommand{\anonym}[1]{#1}

\title{Automatic Detection of Stress from Speech in the Trier Social Stress Test}

\author[affiliation={1}, orcid=0000-0003-3783-7237, correspondingauthor]{Hanna}{Drimalla}
\author[affiliation={1}, orcid=0009-0004-9321-1209, equalcontribution]{Wieland R.}{Cremer}
\author[affiliation={1}, orcid=0000-0001-8751-2399, equalcontribution]{Christine}{Kraus}
\author[affiliation={2}, orcid=0000-0002-9320-2124]{Oliver T.}{Wolf}

\address{
   $^1$ Human-Centered Artificial Intelligence Group, Faculty of Technology, Bielefeld University, Bielefeld, Germany. \\
   $^2$ Department of Cognitive Psychology, Faculty of Psychology, Ruhr University Bochum, Bochum, Germany.
}

\email{drimalla@uni-bielefeld.de}


\usepackage{comment}

\usepackage{booktabs}
\usepackage{makecell}
\usepackage{graphicx}
\usepackage{caption}
\usepackage{subcaption}
\usepackage{siunitx}

\makeatletter
\newcommand{\CurrentFontSize}{\f@size pt}
\makeatother

\begin{document}

\maketitle

\newcommand{\newabbr}[3]{%
  \expandafter\newcommand\csname #1\endcsname{%
    #3\ (#2)%
    \expandafter\gdef\csname #1\endcsname{#2}%
  }%
}

\newabbr{TSST}{TSST}{Trier Social Stress Test}
\newabbr{fTSST}{f-TSST}{friendly-TSST}
\newabbr{sAA}{sAA}{salivary alpha-amylase}
\newabbr{PANAS}{PANAS}{Positive and Negative Affect Schedule}
\newabbr{HPA}{HPA}{hypothalamic–pituitary–adrenal}
\newabbr{PA}{PA}{positive affect}
\newabbr{NA}{NA}{negative affect}
\newabbr{MFCCs}{MFCCs}{Mel-Frequency Cepstral Coefficients}
\newabbr{PCA}{PCA}{principal component analysis}
\newabbr{ML}{ML}{machine learning}
\newabbr{LR}{LR}{logistic regression}
\newabbr{SVM}{SVM}{support vector machine}
\newabbr{SVR}{SVR}{support vector regression}
\newabbr{RF}{RF}{random forest}
\newabbr{RFR}{RFR}{random forest regression}
\newabbr{XGB}{XGB}{XGBoost}
\newabbr{ROC}{ROC}{Receiver Operating Characteristic}
\newabbr{AUC}{AUC}{area under the ROC curve}
\newabbr{LOO}{LOO}{Leave-One-Out}
\newabbr{MAE}{MAE}{mean absolute error}
\newabbr{SHAP}{SHAP}{Shapley Additive Explanations}
\newabbr{Fzero}{F0}{fundamental frequency}
\newabbr{SD}{SD}{standard deviation}

\begin{abstract}
    Automatically detecting stress in speech provides an unobtrusive way to gain insights relevant to behavioral research or clinical assessment. This study investigates the automatic differentiation between a stressful and non-stressful situation, and the prediction of physiological and affective stress responses. Speech data was collected from 50 participants who either completed the Trier Social Stress Test (TSST) or a non-stressful control condition. With a processing pipeline that included speaker diarization and machine learning models, we achieved stress detection performance significantly above a mean baseline. Moreover, relevant physiological and affective stress responses were partially predictable from acoustic-prosodic features. Feature-importance analyses identified the most informative predictors contributing to model performance. The findings demonstrate that speech can serve as a meaningful and unobtrusive indicator of multiple dimensions of the human stress response.
\end{abstract}

\section{Introduction}

In research and clinical practice, stress is commonly measured using self-report instruments and physiological biomarkers such as salivary cortisol and \sAA{}. %
While these measures provide important reference points, they are sensitive to procedural and contextual factors and can be difficult to collect unobtrusively, frequently and at scale in a standardized manner~\cite{brewis2021, bell2021}. These constraints highlight the need for alternative measures that still relate to well-established physiological indices of stress. Speech-based markers represent a particularly promising candidate in this regard. Stress has been linked to systematic modulation of speech production and shown to leave measurable traces in both prosody and voice quality~\cite{giddens2013, vanpuyvelde2018, schewski2025}. Across various stress-inducing settings, speakers show subtle yet reliable acoustic and temporal changes, including shifts in \Fzero{}~\cite{kappen2024Acousticprosodicspeech, pisanski2021}, intensity~\cite{sabo2017designing}, as well as differences in speaking rate or pausing behavior~\cite{pisanski2021, kappen2024Acousticprosodicspeech, buchanan2014}. Moreover, speech can be recorded unobtrusively and repeatedly in both remote and real-world contexts.

Whether speech can serve as a potential biomarker of stress has been investigated via \ML{} on datasets encompassing a broad range of stressors, from speaking in a foreign language~\cite{hanDeep2018} and acted stress~\cite{tomba2018stress}, to experimentally induced stress in laboratory settings~\cite{menne2025Voiceobjectivebiomarker, norden2022automatic}. The variety of stressors and the often unclear presence and intensity of stress may have contributed to inconsistent results~\cite{giddens2013}. Standardized laboratory protocols offer a controlled and well-validated way to experimentally induce stress and link resulting vocal changes to established physiological stress markers. Prior work~\cite{baird2019, baird2021} has investigated such speech-based detection of stress responses using the \TSST{}~\cite{kirschbaum1993}, which is widely considered the ``gold-standard'' among standardized psychosocial stress-inducing paradigms~\cite{dickerson2004}: it reliably elicits acute psychosocial stress through socio-evaluative threat by a speech task and a mental arithmetic task in front of a fake committee. Baird et al.~\cite{baird2019} reported moderate associations between acoustic speech representations and time-varying physiological stress responses measured via salivary cortisol following participation in a \TSST{}. However, to answer the question whether vocal acoustics can be used to distinguish between stressed and non-stressed speech, a control condition is needed. The \fTSST{}~\cite{wiemers2013ftsst} was introduced as such an analogue, preserving the overall structure of the \TSST{} while removing key stress-inducing elements of social-evaluative threat. Recent work incorporating the \fTSST{} has provided first evidence that acoustic cues can distinguish stress and control conditions~\cite{oesten2023}. However, the authors noted an important limitation of their within-subject design: the \fTSST{} may have elicited mild stress responses when the \TSST{} was conducted prior, as reflected in the difficulty of classifying \fTSST{} samples recorded after the \TSST{}. %

In this paper, we therefore aim to evaluate speech-based automated detection of acute psychosocial stress in a fully between-subject setting using a newly collected dataset with both the \TSST{} and the \fTSST{} as a non-stressed control condition. We investigate whether automatically extracted acoustic features %
can (i) discriminate \TSST{} from \fTSST{} speech under cross-participant evaluation, and (ii) predict stress responses indexed by changes in salivary cortisol, \sAA{} and self-reported affect. Further, we explore the most important features contributing to automatic speech-based detection of stress. By combining a matched control condition with affective and physiological outcomes, our work provides further assessment of whether stress-induced speech cues can link to acute stress responses.

\section{Methods}

\subsection{Data Collection}
Data was collected from healthy \anonym{German}-speaking university students participating in a laboratory stress-induction experiment. The study was approved by the local ethics committee of the Faculty of \anonym{Psychology of Ruhr University Bochum} and the Declaration of Helsinki was followed. Exclusion criteria comprised  prior \TSST{} experience; night-shift work; relevant illness, medication use, medical or psychotherapeutic treatment; smoking; substance abuse; and exercising, eating or drinking before testing (see superordinate study~\cite{herten2017} for details). Inclusion criteria included a BMI between 19 and 28 and, for female participants, the use of monophasic oral contraceptives during pill intake to reduce variability in endocrine stress responses related to menstrual-cycle phase. Participants provided informed consent and were randomly assigned to either the stress (\TSST{}) or control (\fTSST{}) condition. In the \textit{stress condition}, a modified version~\cite{wiemers2013modifiedtsst} of the \TSST{}~\cite{kirschbaum1993} was used. Following a \SI{5}{min} preparation phase, participants delivered \SI{10}{min} of free speech framed as a simulated job interview in front of a reserved two-person committee (one male, one female) while being videotaped. The committee did not provide verbal feedback and prolonged pauses were tolerated. In the \textit{control condition} (\fTSST{}~\cite{wiemers2013ftsst}), the same structure was applied without the stress-inducing socio-evaluative threat elements: participants were informed about being in the control condition and could choose from a set of topics; the committee was introduced as laboratory employees to have a friendly conversation with; and sessions were explicitly not videotaped. Unlike in the \TSST{}, the committee %
engaged by asking follow-up questions.
At baseline (\SI{-1}{min}), participants completed the \PANAS{}~\cite{watson1988} and provided a saliva sample. During the preparation phase and the speech task, audio was recorded via eye‑tracking glasses. After the  task, the second saliva sample and \PANAS{} were collected (\SI{+1}{min}). A third saliva sample was obtained at \SI{+20}{min} post‑manipulation.

\subsection{Stress measures}
\subsubsection{Salivary biomarkers}
Saliva samples were analyzed for salivary cortisol, reflecting \HPA{} axis activity~\cite{hellhammer2009}, and \sAA{}, indicating sympathetic nervous system arousal~\cite{nater2009}. Saliva was collected using Salivettes and stored at \SI{-18}{\degreeCelsius} until analysis. Salivary cortisol concentrations (nmol/L) were quantified using a Dissociation-Enhanced Lanthanide Fluorescent Immunoassay (DELFIA~\cite{delfia}; detection limit \SI{0.5}{nmol/L}). \sAA{} activity (U/L) was assessed via a colorimetric test using the CNP-G3 substrate reagent~\cite{lorentz1999saa, winn-deen1988saa}. Biomarker reactivity indices were computed as the maximum post-task value minus the baseline value~\cite{miller2007, khoury2015}, resulting in values for cortisol reactivity and \sAA{} reactivity as ground truth variables.%

\subsubsection{Self-reported affect}
\PANAS{}~\cite{watson1988} is a validated 20-item measure of positive and negative affect. Participants indicate the intensity of 10 positive and 10 negative emotions on a 5-point Likert scale (1 = ‘very slightly or not at all’,  5 = ‘extremely’). Positive (PA) and negative affect (NA) scores were obtained for pre- and post-manipulation. Change scores ($\Delta$PA and $\Delta$NA; post $-$ pre) served as ground truth variables.

\subsection{Audio data}
Audio was recorded via the built-in microphone of SMI Eye Tracking Glasses 2.0 \anonym{(SensoMotoric Instruments GmbH, Teltow, Germany)} %
as uncompressed \SI{16}{kHz} \textit{.wav} files. Recordings started at the onset of the preparation phase and ended shortly after task completion. To reduce irrelevant noise (e.g., experimenter interaction), all raw audio files (\TSST{}: 16.62 \(\pm\) 0.32~min; \fTSST{}: 16.72 \(\pm\) 0.34~min) were trimmed to \SI{9}{min} segments starting at minute 7. 
For isolating participant speech, speech diarization was performed automatically using \textit{Sortformer}~\cite{park2025sortformer}, a pretrained transformer encoder-based end-to-end speaker diarization model by NVIDIA NeMo Speech AI. The participant was identified as the speaker with the longest total speaking time. Participant-only audio was generated by retaining their diarized speech segments, removing overlaps (\SI{50}{ms} collar) and concatenating the segments into a single waveform per recording without pauses longer than those occurring in natural speech (mean lengths of recordings after processing: \TSST{}: 4.13 \(\pm\) 1.85~min; \fTSST{}: 6.51 \(\pm\) 0.98~min). A random subset ($n = 12$) of the pre-processed recordings was manually inspected to assess diarization quality, ensure the absence of non-participant speech and compared to diarization using \textit{pyannote}~\cite{bredin23, plaquet23}. Noise accounted for less than 5\% of the total recording time for each participant. Acoustic features were then extracted from the participant speech using three complementary toolchains: \textit{librosa}~(v0.11.0)~\cite{mcfee2015librosa} was used to extract 40 \MFCCs{}. Audio was resampled to \SI{22.05}{kHz} and \MFCCs{} were computed frame-wise and then averaged. Furthermore, 15 classical voice parameters (e.g., mean/SD \Fzero{}, HNR, median pitch, jitter, shimmer) were extracted using \textit{Praat}~(v6.1.38)~\cite{praat} via Parselmouth~(v0.4.7)~\cite{parselmouth}. Additionally, the eGeMAPSv02 feature set~\cite{eyben2016egemaps} was extracted using \textit{openSMILE}~(v2.6.0)~\cite{eyben2010opensmile}, containing 88 statistical functionals summarizing pitch, energy, spectral and other voice-quality measures. For each participant, feature sets were concatenated into a single participant-level vector with sex added as a covariate to account for related differences in the voice, resulting in a 144‑dimensional feature vector per participant. Within each cross-validation split, each feature was $z$-standardized across participants in the training fold. The same  $z$-standardization was  subsequently applied to the corresponding hold-out participant-level feature vectors.

\subsection{Machine learning models and evaluation}
All models were trained on the participant-level feature vectors obtained from the audio recordings. For comparison, models were additionally trained on reduced-dimensional representations obtained via \PCA{}. The code for preprocessing, \ML{} analysis and evaluation as well as additional figures are publicly available on GitHub %
(\url{https://github.com/mbp-lab/tsst-speech-stress}). 

\subsubsection{Classification}
The objective of the binary classification task was to distinguish between participants who underwent the \TSST{} and those who underwent the \fTSST{}. Four classification algorithms, covering a range of model complexities from linear to nonlinear ensemble methods, were trained: \LR{}~\cite{cox1958logisticregression}, \SVM{}~\cite{cortes1995svm}, \RF{} classifier~\cite{breiman2001randomforest} and \XGB{} classifier~\cite{chen2016xgboost}. The \LR{} model was tuned for the regularization coefficient  $\lambda \in \{0.1, 1, 2, 10, 100\}$ and penalty type (L1 or L2). The \SVM{} was optimized with respect to the kernel function (linear or RBF), the regularization coefficient $\lambda \in \{0.01, 0.1, 1, 10\}$ and for the RBF kernel, the kernel coefficient $\gamma \in \{\texttt{scale}, 0.001, 0.01, 0.1, 1, 10\}$. For \RF{}, the number of trees was fixed at 1000, while maximum tree depth $\{1, 2, 4, 8\}$ and minimum number of samples per split $\{1, 2, 4\}$ were tuned. For \XGB{}, the number of trees $\{50, 100, 150\}$, maximum tree depth $\{1, 2, 4, 8\}$ and learning rate $\{0.03, 0.1, 0.2\}$ were optimized.

\subsubsection{Regression}
Regression analyses were conducted to predict different stress responses from speech-derived acoustic features. Ground-truth variables included cortisol and \sAA{} reactivity and changes in positive and negative affect: for each, separate models were calculated. Three regression algorithms were trained: \SVR{}~\cite{drucker1996svr}, a \RFR{}~\cite{breiman2001randomforest} and an \XGB{} regressor. The \SVR{} was tuned over the kernel function (linear or RBF), the regularization coefficient $\lambda \in \{0.01, 0.1, 1, 10\}$ and for the RBF kernel, the kernel coefficient 
$\gamma \in \{\texttt{scale}, 0.001, 0.01, 0.1, 1, 10\}$. For \RFR{}, the number of trees was fixed at 1000, while the maximum tree depth $\{2, 4, 5, 10\}$ was optimized. For the \XGB{} regressor, number of trees $\{50, 100, 150\}$, maximum tree depth $\{1, 2, 4, 8\}$ and learning rate $\{0.03, 0.1, 0.2\}$ were optimized. Separate models were trained on the full sample and the \TSST{} subsample.

\subsubsection{Evaluation}

Cross-validation was performed across participants, with each participant represented by a single feature vector. All preprocessing steps, including feature standardization and \PCA{}, were carried out exclusively within the training folds. For classification, a nested cross-validation scheme was used, with an outer 10-fold and an inner 3-fold cross-validation for hyperparameter tuning. Performance was evaluated using classification accuracy and the \AUC{}. We included a majority-class baseline and conducted corrected paired $t$-test proposed by Nadeau and Bengio~\cite{nadeau2003ttest} that accounts for the dependency due to the cross-validation scheme. For regression, a nested \LOO{} cross-validation scheme was used, with an outer \LOO{} loop and an inner 5-fold cross-validation for hyperparameter tuning. Performance is reported as \MAE{} and Spearman's correlation between true and predicted values. A mean-value baseline \MAE{} was calculated on each fold, averaged across folds and tested for statistical significance using corrected paired $t$-test by Nadeau and Bengio~\cite{nadeau2003ttest}. Because overlapping training folds violate the independence assumption, significance tests are not reported for correlations. Feature importance was computed through \SHAP{}~\cite{lundberg2017shap} and averaged across folds.

\section{Results}

\subsection{Participants}
After exclusions (five dropouts, five instances of data loss, one corrupted data recording, one baseline cortisol outlier $>$ 3 \SD{} and one cortisol non-responder), the final sample comprised 50 participants (25 per condition). Overall, mean age was 23.24 years (\(\pm\)3.92) and mean BMI was 23.28 (\(\pm\)2.35); 23 (46.0\%) participants were female. Descriptives by condition were comparable (\TSST{}: mean age 22.48 \(\pm\)3.45, mean BMI 23.30 \(\pm\)2.77, 12 female (48.0\%); \fTSST{}: mean age 24 (\(\pm\)4.27), mean BMI 23.27 (\(\pm\)1.90), 11 female (44.0\%)).

\subsection{Manipulation check}
To verify successful stress induction, we compared physiological and affective responses between \TSST{} and \fTSST{}. Log-transformed salivary cortisol and \sAA{} reactivity indices, along with changes in self-reported affect ($\Delta$NA, $\Delta$PA), were analyzed. Cortisol responses were higher in the \TSST{} group, with mixed-effects models showing significantly greater increases at \SI{+1}{min} ($b = 0.40$, $p = .001$) and \SI{+20}{min} ($b = 0.65$, $p < .001$), and no baseline difference ($b = 0.18$, $p = .27$). The \sAA{} trajectories did not differ significantly between conditions (\SI{+1}{min}: $b = -0.02$, $p = .85$; \SI{+20}{min}: $b = -0.03$, $p = .78$). Negative affect increased under stress and decreased in the control condition ($\Delta$NA: \TSST{} $3.04 \pm 4.04$; \fTSST{} $-2.12 \pm 4.58$; $b = 3.96$, $p = .019$), while positive affect showed the opposite, nonsignificant trend ($\Delta$PA: \TSST{} $-0.04 \pm 4.68$; \fTSST{} $4.28 \pm 5.42$; $b = -3.72$, $p = .065$).

\subsection{Classifications}

\begin{figure}[t]
    \centering
    \includegraphics[width=\columnwidth]{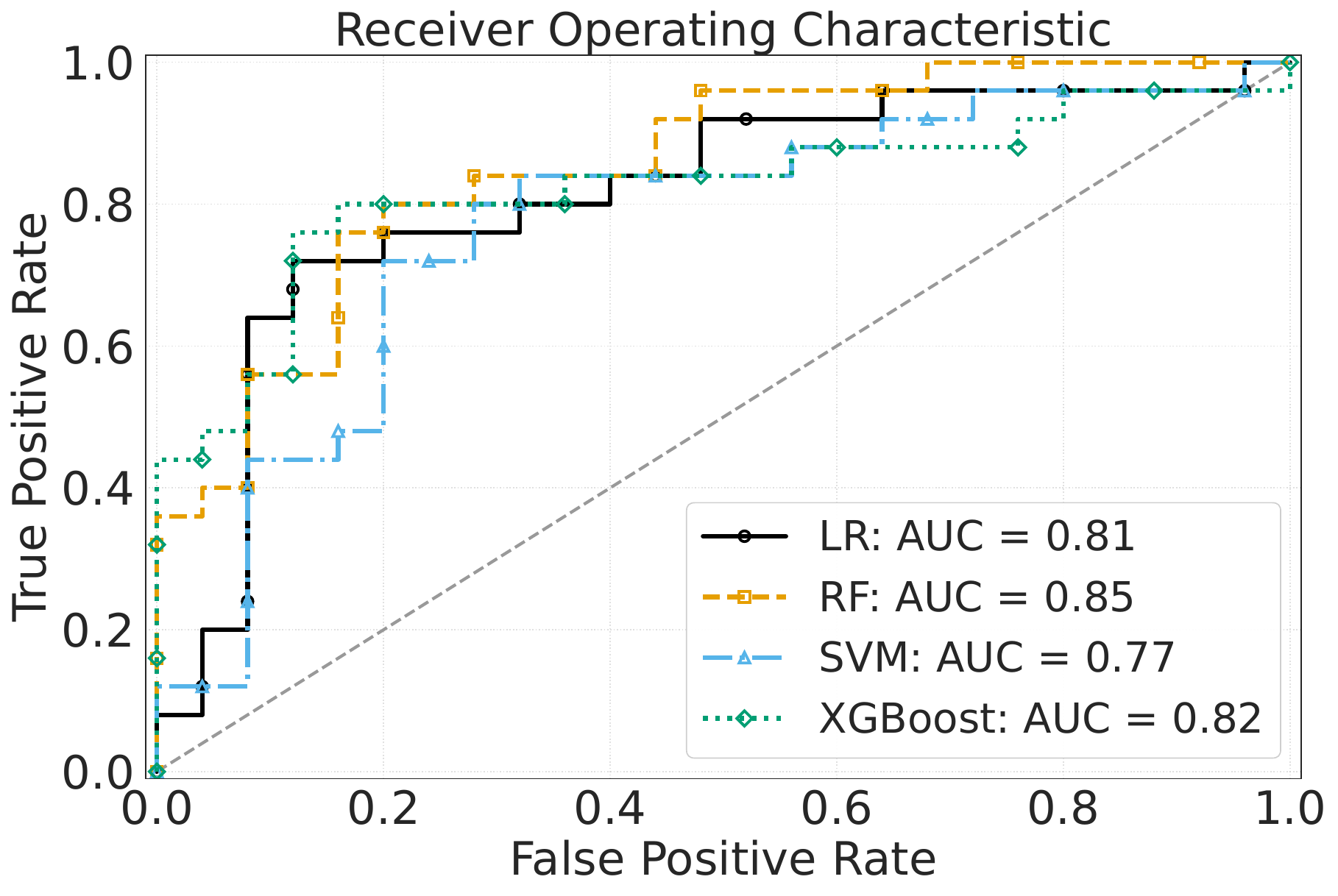}
    \caption{\ROC{} curves for the classification models.}
    \label{fig:roc}
\end{figure}

\begin{figure}[t]
    \centering
    \includegraphics[width=\columnwidth]{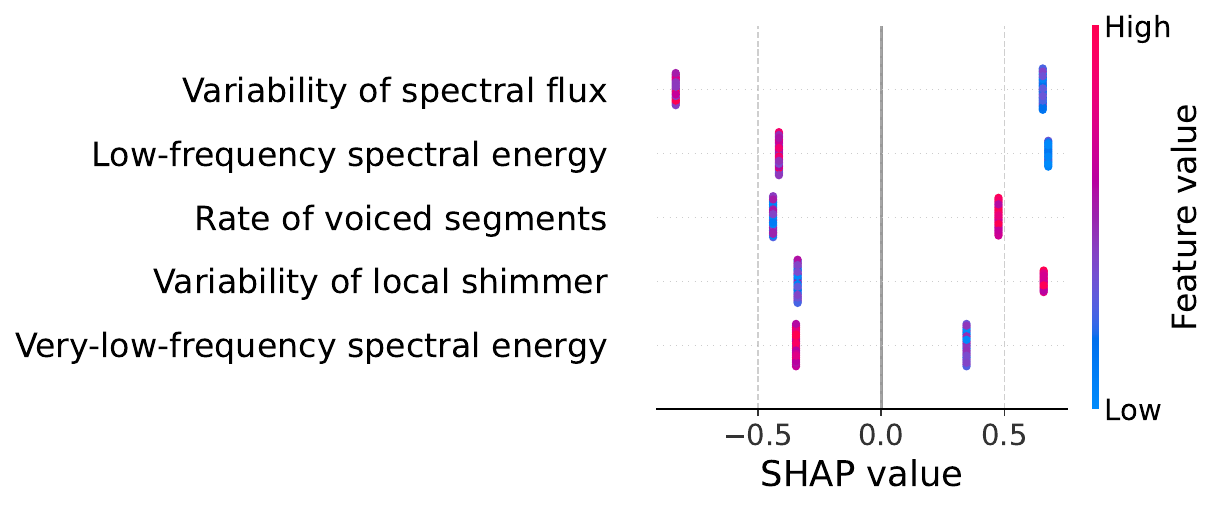}
    \caption{Top 5 \SHAP{} values for \XGB{} Classifier.}
    \label{fig:shap}
\end{figure}

The four \ML{} models classified whether a participant was in the stress condition (\TSST{}) or the control condition (\fTSST{}) %
. The highest accuracy was achieved by the \XGB{} classifier (accuracy $0.82 \pm 0.11$ %
), outperforming the majority-class baseline (corrected $t = 8.05$, $p < .001$). Misclassifications were relatively balanced, as illustrated by the confusion matrix in Table~\ref{tab:confusionmatrix}. Based on \SHAP{} values averaged across all folds, the most relevant features were identified as the variability of voiced spectral flux, very‑low‑ and low‑frequency spectral energy, the rate of voiced speech segments and the variability of local shimmer (see Figure~\ref{fig:shap}). The \RF{} showed comparable performance (accuracy $0.80 \pm 0.18$%
; corrected $t = 4.62$, $p = .001$). \LR{} reached an accuracy of $0.78 \pm 0.23$ %
(corrected $t = 3.45$, $p = .004$), while \SVM{} achieved an accuracy of $0.74 \pm 0.18$ %
; corrected $t = 3.90 $, $p = .002$. The corresponding \ROC{} curves are shown in Figure~\ref{fig:roc}. Additional model variants using dimensionality reduction with \PCA{} did not yield improved performance.

\begin{table}[t]
\centering
\caption{Confusion matrix for the \XGB{} classifier.}
\label{tab:confusionmatrix}
\setlength{\tabcolsep}{6pt}
\renewcommand{\arraystretch}{1.2}
\begin{tabular}{lcc}
\toprule
& \textbf{Predicted \TSST{}} & \textbf{Predicted \fTSST{}} \\
\midrule
\textbf{Actual \TSST{}}   & 20 & 5  \\
\textbf{Actual \fTSST{}} & 4  & 21 \\
\bottomrule
\end{tabular}
\end{table}

\subsection{Regressions}

\begin{table*}[!t]
\centering
\caption{Regression results comparison.}
\label{tab:reg_all}

\textbf{(A) Performance of regression models for predicting stress responses (full sample)}\par
\vspace{0.2em}
\resizebox{\textwidth}{!}{%

\begin{tabular}{l c c c c c c c c c c c c c c c}
  \toprule
  \multicolumn{1}{c}{\textbf{Model}} &
  \multicolumn{6}{c}{\textbf{Cortisol}} &
  \multicolumn{3}{c}{\textbf{sAA}} &
  \multicolumn{6}{c}{\textbf{PANAS (Affect)}} \\
  \cmidrule(lr){2-7}\cmidrule(lr){8-10}\cmidrule(lr){11-16}

  & \multicolumn{3}{c}{\textbf{Reactivity}} &
    \multicolumn{3}{c}{\textbf{+20 min}} &
    \multicolumn{3}{c}{\textbf{Reactivity}} &
    \multicolumn{3}{c}{\textbf{$\Delta$NA}} &
    \multicolumn{3}{c}{\textbf{$\Delta$PA}} \\
  \cmidrule(lr){2-4}\cmidrule(lr){5-7}\cmidrule(lr){8-10}\cmidrule(lr){11-13}\cmidrule(lr){14-16}

  & \textbf{MAE} & \makecell{\textbf{t ($p$)}} & \makecell{\textbf{$\rho$}} &
    \textbf{MAE} & \makecell{\textbf{t ($p$)}} & \makecell{\textbf{$\rho$}} &
    \textbf{MAE} & \makecell{\textbf{t ($p$)}} & \makecell{\textbf{$\rho$}} &
    \textbf{MAE} & \makecell{\textbf{t ($p$)}} & \makecell{\textbf{$\rho$}} &
    \textbf{MAE} & \makecell{\textbf{t ($p$)}} & \makecell{\textbf{$\rho$}} \\
  \midrule

  RFR &
  \textbf{3.73} & 0.68 (.25) & 0.20 &
  5.04 & -0.22 (.41) & 0.10 &
  \textbf{32.37} & 1.04 (.15) & 0.21 &
  3.17 & -0.09 (.47) & 0.17 &
  4.11 & -0.50 (.31) & -0.10 \\

  SVR &
  \textbf{3.10} & 2.01 (.02) & 0.01 &
  \textbf{4.78} & 0.49 (.32) & -0.11 &
  \textbf{35.87} & 0.77 (.22) & -0.82 &
  3.37 & -0.57 (.29) & 0.28 &
  3.96 & -0.17 (.43) & 0.05 \\

  XGB &
  \textbf{3.41} & 0.98 (.17) & 0.42 &
  5.63 & -1.13 (.13) & -0.08 &
  \textbf{35.55} & 0.27 (.39) & 0.18 &
  \textbf{3.10} & 0.07 (.47) & 0.49 &
  4.15 & -0.58 (.28) & -0.13 \\

  Dummy &
  4.04 & {--} & {--} &
  4.93 & {--} & {--} &
  37.27 & {--} & {--} &
  3.14 & {--} & {--} &
  3.93 & {--} & {--} \\
  \bottomrule
\end{tabular}

}

\vspace{0.5em}

\textbf{(B) Performance of regression models for predicting stress responses (\TSST{} subsample)}\par
\vspace{0.2em}
\resizebox{\textwidth}{!}{%

\begin{tabular}{l c c c c c c c c c c c c c c c}
  \toprule
  \multicolumn{1}{c}{\textbf{Model}} &
  \multicolumn{6}{c}{\textbf{Cortisol}} &
  \multicolumn{3}{c}{\textbf{sAA}} &
  \multicolumn{6}{c}{\textbf{PANAS (Affect)}} \\
  \cmidrule(lr){2-7}\cmidrule(lr){8-10}\cmidrule(lr){11-16}

  & \multicolumn{3}{c}{\textbf{Reactivity}} &
    \multicolumn{3}{c}{\textbf{+20 min}} &
    \multicolumn{3}{c}{\textbf{Reactivity}} &
    \multicolumn{3}{c}{\textbf{$\Delta$NA}} &
    \multicolumn{3}{c}{\textbf{$\Delta$PA}} \\
  \cmidrule(lr){2-4}\cmidrule(lr){5-7}\cmidrule(lr){8-10}\cmidrule(lr){11-13}\cmidrule(lr){14-16}

  & \textbf{MAE} & \makecell{\textbf{t ($p$)}} & \textbf{$\rho$} &
    \textbf{MAE} & \makecell{\textbf{t ($p$)}} & \textbf{$\rho$} &
    \textbf{MAE} & \makecell{\textbf{t ($p$)}} & \textbf{$\rho$} &
    \textbf{MAE} & \makecell{\textbf{t ($p$)}} & \textbf{$\rho$} &
    \textbf{MAE} & \makecell{\textbf{t ($p$)}} & \textbf{$\rho$} \\
  \midrule

  RFR &
  \textbf{4.93} & 0.73 (.24) & 0.27 &
  5.81 & -0.35 (.37) & 0.09 &
  \textbf{40.14} & 0.37 (.36) & 0.18 &
  \textbf{2.82} & 0.94 (.18) & 0.53 &
  3.79 & -0.16 (.44) & -0.06 \\

  SVR &
  \textbf{4.43} & 1.48 (.08) & 0.34 &
  6.20 & -1.30 (.10) & -0.54 &
  43.43 & -0.21 (.42) & 0.07 &
  3.45 & -0.51 (.30) & 0.10 &
  3.96 & -0.69 (.25) & -0.45 \\

  XGB &
  5.55 & 0.00 (.50) & 0.06 &
  5.78 & -0.39 (.35) & -0.08 &
  50.05 & -0.79 (.22) & 0.12 &
  \textbf{2.08} & 2.11 (.02) & 0.67 &
  3.74 & -0.04 (.48) & -0.01 \\

  Dummy &
  5.55 & {--} & -- &
  5.53 & {--} & -- &
  42.34 & {--} & -- &
  3.22 & {--} & -- &
  3.72 & {--} & -- \\
  \bottomrule
\end{tabular}

}

\vspace{0.5em}
{\fontsize{8}{11}\selectfont\textit{Note.} Dummy =  mean baseline. $t(p)$ = corrected $t$-test: outperforms dummy. $\rho$ = Spearman’s correlation. Boldface = lower MAE than baseline.}

\end{table*}

\ML{} regression models were trained to predict both physiological and affective stress reactivity. %
Among the stress responses, cortisol reactivity and negative affect could be predicted from speech-derived features, with performance varying across models (Table~\ref{tab:reg_all}). Using the full dataset, including participants from both the stress and control conditions, the \SVR{} model predicted cortisol reactivity more accurately than the baseline. The most relevant features for this prediction were the rate of voiced segments, low and mid‑frequency spectral energy, variation in spectral tilt and the spread of low pitch values. For the \TSST{} subsample, the \SVR{} achieved a lower \MAE{} than the mean baseline, although the improvement was only marginally significant. %
For negative affect, the \XGB{} regressor outperformed the baseline, but the improvement reached statistical significance only for the \TSST{} subsample. The most relevant acoustic features were the mean \Fzero{} rising slope, F1 bandwidth, Hammarberg index, %
alpha ratio, %
and the \SD{} of \Fzero{}.

\section{Discussion}

Our work demonstrates that subtle acoustic-prosodic characteristics of speech can be leveraged for automatic stress detection. The randomized between-participant design established a validated \TSST{}/\fTSST{} contrast in stress, as confirmed by the manipulation check. The experimental conditions could be automatically distinguished from speech-derived features using different \ML{} models; although influence of protocol-related cues cannot be fully excluded, the successful stress manipulation supports interpreting the learned acoustic differences as stress-related.%
This extends not only studies without a control condition~\cite{baird2019, baird2021} but also a  within‑subject study~\cite{oesten2023}, where order effects may have confounded results. %
Furthermore, we were able to predict key markers of distinct stress responses elicited by the \TSST{}, namely cortisol reactivity and changes in negative affect, with performance exceeding that of a mean‑baseline regressor. Statistical testing confirmed these effects, at least for the best‑performing models, for relevant physiological and affective stress indices. Also, predicted values showed positive correlations with observed responses.
In line with previous research, the feature-importance analysis identified features related to pitch~\cite{kappen2024Acousticprosodicspeech, pisanski2021}, speech productivity~\cite{buchanan2014, pisanski2021, kappen2024Acousticprosodicspeech} and shimmer~\cite{park2011, kappen2022}. In addition, some less frequently studied features emerged, which have recently shown promising associations with stress, namely the alpha ratio~\cite{menne2025Voiceobjectivebiomarker} and the Hammarberg index~\cite{tavi2017Acousticcorrelatesfemale}.

In contrast to related work~\cite{baird2019, baird2021}, it was not possible to consistently predict the \SI{+20}{min} cortisol value. This limitation may stem from not standardizing these measurements due to the small number of available time‑points. However, since cortisol reactivity is considered a key marker of physiological stress~\cite{miller2007, khoury2015}, we regard its successful prediction especially relevant. That \sAA{} reactivity could not be predicted is unsurprising, given that, consistent with previous work~\cite{wiemers2013ftsst}, the \fTSST{} also induces a \sAA{} response.
Beyond the physiological response, we also predicted the negative affect change, which is important given that previous work~\cite{norden2022automatic} has shown the necessity of considering multiple facets of stress in automatic stress‑response modeling.
While differences in committee interaction between \TSST{} and \fTSST{} limit direct comparability of pause structures, the same pattern for cortisol reactivity and $\Delta$NA observed in \TSST{}-only regressions supports the robustness of our results.
With \XGB{}, we chose a classifier well-suited for feature-level interpretation and that has been shown to outperform deep learning methods on tabular data~\cite{shwartz2022tabular}. Nevertheless, future research should compare such interpretable feature-based approaches with modern pretrained, end-to-end and multimodal deep learning models in similarly controlled designs, particularly as architectures incorporating temporal information, such as LSTMs, have been shown to improve performance~\cite{baird2021}. %
Overall, our findings highlight speech as a promising digital biomarker of stress and demonstrate the feasibility of automated processing for stress detection and both physiological and affective stress response prediction.
The analysis pipeline may serve as a useful basis for advancing objective stress assessment in both research and clinical practice.

\section{Acknowledgments}
\anonym{The project was funded by the Deutsche Forschungsgemeinschaft (DFG, German Research Foundation): TRR 318/3 2026 -- 438445824. The superordinate study~\cite{herten2017} that provided the data was supported by the DFG project B4 of the Collaborative Research Centre (SFB) 874 ``Integration and Representation of Sensory Processes'' awarded to Oliver T. Wolf. 
We would like to thank Dennis Pomrehn, Nadja Herten and Sarah Weusthoff for their valuable contributions to this work during their time at the Department of Cognitive Psychology, Faculty of Psychology, Ruhr University Bochum.}

\section{Generative AI Use Disclosure}
Generative AI was used only for editing and polishing the manuscript.

\bibliographystyle{IEEEtran}
\bibliography{mybib}

\end{document}